\pdfoutput=1
\documentclass[10pt,twocolumn,letterpaper]{article}

\usepackage[pagenumbers]{cvpr}
\usepackage{algorithm}
\usepackage{algorithmic}
\usepackage{amsfonts}
\usepackage{array}
\usepackage{colortbl}
\usepackage{multirow}
\usepackage{newfloat}
\usepackage{pifont}
\usepackage{pgfplots}
\usepackage{tabularx}
\usepackage{listings}

\pgfplotsset{compat=1.18}

\newcolumntype{C}{>{\centering\arraybackslash}X}

\DeclareCaptionStyle{ruled}{labelfont=normalfont,labelsep=colon,strut=off}
\floatstyle{ruled}
\newfloat{listing}{tb}{lst}{}
\floatname{listing}{Listing}

\definecolor{cvprblue}{rgb}{0.21,0.49,0.74}
\usepackage[breaklinks,colorlinks,allcolors=cvprblue]{hyperref}

\def\confName{CVPR}
\def\confYear{2027}

\title{CoDy: Codec Dynamics as Complementary Priors for Unsupervised Temporal Action Localization}

\author{Tianyu Yan\textsuperscript{1}\quad
Qingyi Yang\textsuperscript{1}\quad
Qin Yang\textsuperscript{1}\quad
Xufei Guo\textsuperscript{1}\\
Yuan Tang\textsuperscript{2}\quad
Rui Wang\textsuperscript{1}\quad
Yixue Hao\textsuperscript{1}\quad
Long Hu\textsuperscript{1}\\[0.5em]
\textsuperscript{1}School of Computer Science and Technology,\\
Huazhong University of Science and Technology\\
\textsuperscript{2}Huawei Technologies Ltd\\
{\tt\small tianyuyan@hust.edu.cn}
}

\begin{document}

\maketitle

\begin{abstract}
Unsupervised temporal action localization (UTAL) aims to identify action categories and localize their temporal boundaries without annotation supervision.
Existing methods typically derive supervision for both tasks solely from RGB and optical-flow representations, often resulting in noisy category assignments and inaccurate temporal boundaries.
In this work, we reveal an overlooked property of video codecs: group-of-pictures quantization patterns preserve category-discriminative dynamics, while motion-vector and residual variations exhibit pronounced responses around action transitions.
Based on this insight, we propose CoDy, a codec dynamics-guided framework that exploits the codec as complementary information to the visual view for UTAL.
CoDy integrates a codec-enhanced view with RGB and flow through confidence-aware consensus clustering, producing reliable soft pseudo-labels.
It further converts codec dynamics into a class-agnostic boundary prior, which supervises visual boundary learning and guides proposal initialization and iterative refinement.
The codec branches are only used during the training stage and introduce no extra codec-side computation at inference.
Extensive experiments on THUMOS14 and ActivityNet v1.3 demonstrate that CoDy consistently outperforms existing UTAL methods and achieves state-of-the-art localization performance.
These results establish codec signals as effective annotation-free priors for jointly inferring which actions occur and when they occur.
\end{abstract}

\section{Introduction}
\label{sec:intro}
Temporal action localization (TAL) aims to identify the actions occurring in an untrimmed video and determine the temporal boundaries of each action instance~\cite{shou2016temporal,zhao2017structured}.
%
Fully supervised methods learn this task from action-category labels and temporal annotations~\cite{xu2017r3c3d,zhao2017structured}, but collecting temporal boundary annotations is expensive and inherently ambiguous~\cite{moltisanti2019action}.
%
Weakly supervised TAL reduces the required supervision to video-level action labels~\cite{nguyen2018stpn,zhang2021cola,ren2023pmil}, whereas unsupervised temporal action localization (UTAL) removes these labels entirely and assumes only that the number of latent action categories is known~\cite{gong2020tcam,yang2023ugct}.
%
Without semantic or temporal annotations, UTAL needs to jointly discover latent action categories and localize their temporal extents. These two objectives are inherently coupled: category discovery requires foreground representations that are not dominated by background or incomplete action segments, while accurate localization requires reliable category semantics to distinguish actions from their surrounding context. 
Thus, there is a coupled challenge in UTAL between discovery and localization, in which errors from either side can be amplified.

\begin{figure}
  \centering
  \includegraphics[width=\columnwidth]{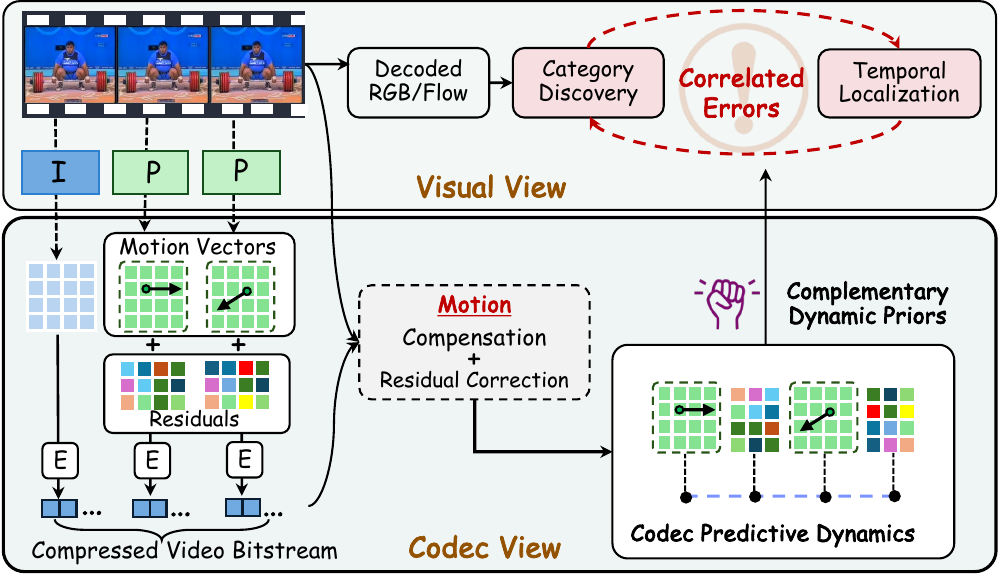}
  \caption{The codec view's predictive dynamics provide complementary priors to correct errors in the visual view.}
  \label{fig:motivation}
\end{figure}

Existing UTAL methods typically address this problem through an iterative paradigm. Visual representations are iteratively clustered to generate pseudo-category labels, which guide temporal localization and the resulting action proposals are then fed back to refine the pseudo-category assignments for the next iteration~\cite{liu2023apsl}.
Recent methods improve this process through cross-view collaboration \cite{xia2025clipae}, uncertainty-aware learning \cite{yang2023ugct}, and pseudo-label refinement \cite{hu2026feel}.
%
Nevertheless, both latent category discovery and temporal boundary estimation still rely predominantly on decoded RGB and optical-flow representations.
\textbf{For action category discovery,} visual view captures appearance and apparent motion. However, RGB features often entangle actions with scene context, while optical flow is vulnerable to camera motion, weakening the category structure required for clustering.
\textbf{For action temporal localization,} learned representations tend to emphasize discriminative action interiors rather than temporal transitions, leading to incomplete or over-extended boundaries.
Consequently, the two stages inherit correlated errors from the same visual evidence, which may be reinforced rather than corrected during iteration. It requires complementary evidence to break this loop that can recognize latent actions while exposing their temporal transitions.


We find that the predictive structure of video coding provides a promising source of such evidence \cite{legall1991mpeg,wiegand2003h264,wu2018coviar}.
A video codec organizes frames into groups of pictures (GOPs) and represents temporal evolution through motion-compensated prediction and residual correction. Motion vectors describe predictable displacement, while quantized residuals retain variations unexplained by prediction~\cite{legall1991mpeg,wiegand2003h264}. 
RGB and optical flow describe the observed visual states, whereas codec signals record how these states are predicted and corrected over time, providing a distinct view of temporal dynamics.
Previous compressed-domain studies have demonstrated that codec primitives carry learnable video information~\cite{wu2018coviar,chen2022mmvit,sarkar2026cope}. 
In contrast, we investigate whether the predictive structure preserved by video coding can serve as an intrinsic prior for UTAL.
Our empirical analysis reveals two complementary properties: GOP-level quantization patterns preserve category-discriminative dynamics complementary to RGB and flow, while variations in motion vectors and quantized residuals exhibit pronounced responses around action transitions. 
Although these signals do not directly specify action categories or exact temporal boundaries, they provide annotation-free priors for discovering {what} actions occur and determining {when} they occur.


Building on these findings, we propose \textbf{CoDy}, a codec dynamics-guided framework that complements decoded visual representations with predictive dynamics for UTAL. 
As illustrated in Figure~\ref{fig:motivation}, RGB and flow provide the primary visual evidence for action discovery and localization, while codec signals introduce complementary structural priors. 
For category discovery, CoDy constructs a GOP-level codec dynamics view. It aligns it with RGB and optical flow through confidence-aware multi-view consensus clustering, yielding reliable soft pseudo-labels. 
Based on these labels, the motion-vector and quantized-residual variations provide class-agnostic boundary priors for multi-scale temporal refinement. 
Codec-derived proposals further optimize the loop with more reliable action candidates, which are subsequently fed back to refine the multi-view category assignments.
Codec information is required only during training and inference retains the learned localization network and boundary branch without additional codec processing. Extensive experiments on THUMOS14 and ActivityNet v1.3 establish new state-of-the-art performance, with particularly pronounced gains at high IoU thresholds. 
Our contributions are summarized as follows:
\begin{itemize}
    \item We revisit predictive video coding and empirically reveal that GOP-level quantization preserves category-discriminative dynamics, while motion-vector and residual variations provide cues to action transitions.
    \item We propose CoDy, a codec-prior UTAL framework that translates these complementary structures into pseudo-label generation and boundary prior learning, thereby stabilizing the coupled discovery and localization.
    \item Extensive experiments on THUMOS14 and ActivityNet v1.3 establish new state-of-the-art performance, with pronounced improvements at high IoU thresholds and validate the effectiveness and generality of codec priors.
\end{itemize}

\begin{figure*}[t]
  \centering

  \begin{subfigure}[t]{0.65\textwidth}
    \centering
    \includegraphics[width=\linewidth]{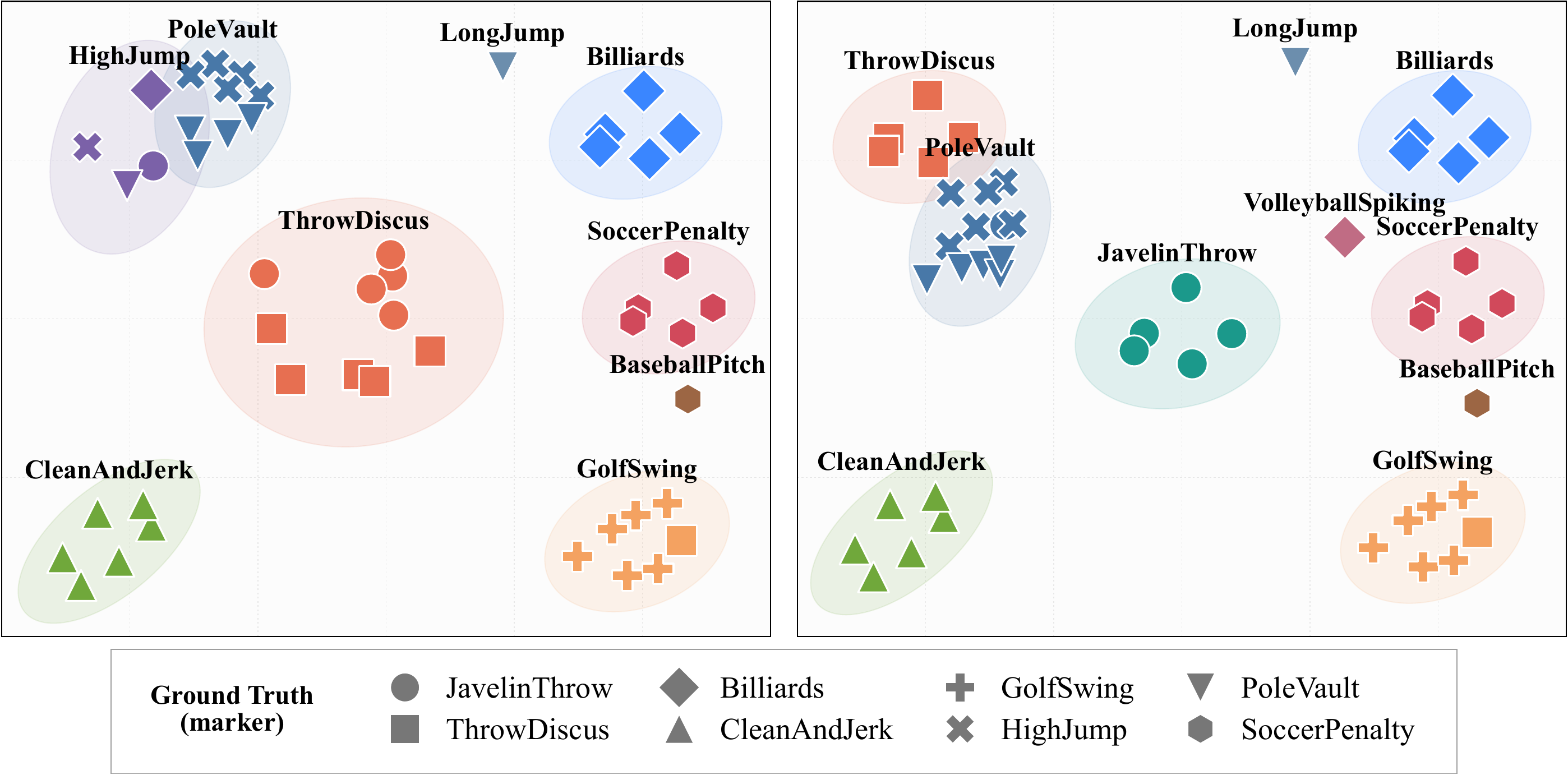}
    \caption{Category clustering based on visual view (Left) and complementary codec view (Right).}
    \label{fig:sec3_1}
  \end{subfigure}
  \hfill
  \begin{subfigure}[t]{0.345\textwidth}
    \centering
    \includegraphics[width=\linewidth]{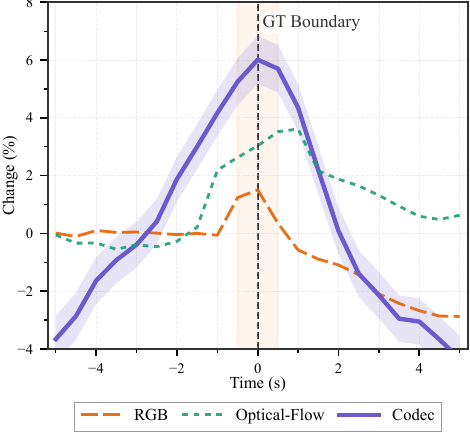}
    \caption{Response-boundary relationship.}
    \label{fig:sec3_2}
  \end{subfigure}

  \caption{Empirical results of codec-prior dynamics for category discrimination and boundary sensitivity detection.}
  \label{fig:sec3}
\end{figure*}

\section{Related Work}
\label{sec:related}

\subsection{Unsupervised Temporal Action Localization}

UTAL aims to discover latent action categories and localize their temporal intervals in untrimmed videos without any human-provided annotations. TCAM~\cite{gong2020tcam} establishes the clustering-and-updating paradigm for UTAL, where videos are iteratively clustered into latent categories and the resulting assignments are used as pseudo supervision for localization learning. Subsequent works have improved this framework from different perspectives. UGCT~\cite{yang2023ugct} leverages RGB--flow collaboration and uncertainty estimation to suppress noisy pseudo supervision. APSL~\cite{liu2023apsl} focuses on separating action-positive evidence from background content, while FEEL~\cite{hu2026feel} improves category discovery through enhanced clustering and self-paced instance selection. More recently, CLIP-AE incorporates visual-language and audio modalities as complementary views for action discovery~\cite{xia2025clipae}. BID explores boundary--interior structural learning for skeleton-based unsupervised temporal action pre-training~\cite{fang2024bid}. Most UTAL methods iteratively cluster video or segment representations to infer latent categories, then use the resulting pseudo-labels to train a localization model and update the representations for subsequent clustering. However, background content, intra-class variation, and similar actions can yield noisy category assignments, while unreliable pseudo-labels provide weak localization supervision. These errors may accumulate across iterations, making robust category discovery and boundary localization the key challenges of UTAL.

\subsection{Codec Representations for Video Understanding}

Modern video codecs reduce temporal and spatial redundancy by encoding videos with intra-coded frames, motion-compensated prediction, and residual signals~\cite{legall1991mpeg,wiegand2003h264}. Specifically, I-frames provide appearance references, motion vectors describe block-wise displacement for inter-frame prediction, and residuals preserve information that cannot be explained by the predicted content. These codec primitives are generated during video compression and can be accessed without fully decoding all RGB frames. Early works exploit compressed-domain representations primarily for efficient video recognition. CoViAR demonstrates that I-frames, motion vectors, and residuals can serve as complementary inputs for action recognition, avoiding costly dense RGB decoding~\cite{wu2018coviar}. Later, MM-ViT models interactions among these heterogeneous codec signals with a compressed-video transformer~\cite{chen2022mmvit}, while CoPE-VideoLM represents codec primitives as compact visual tokens for video-language modeling~\cite{sarkar2026cope}. Overall, existing studies mainly treat codec information as an efficient alternative or complement to decoded visual features.

\section{Empirical Analysis of Codec-Prior Dynamics}
\label{sec:observations}

\begin{figure*}[t]
  \centering
  \includegraphics[width = \textwidth]{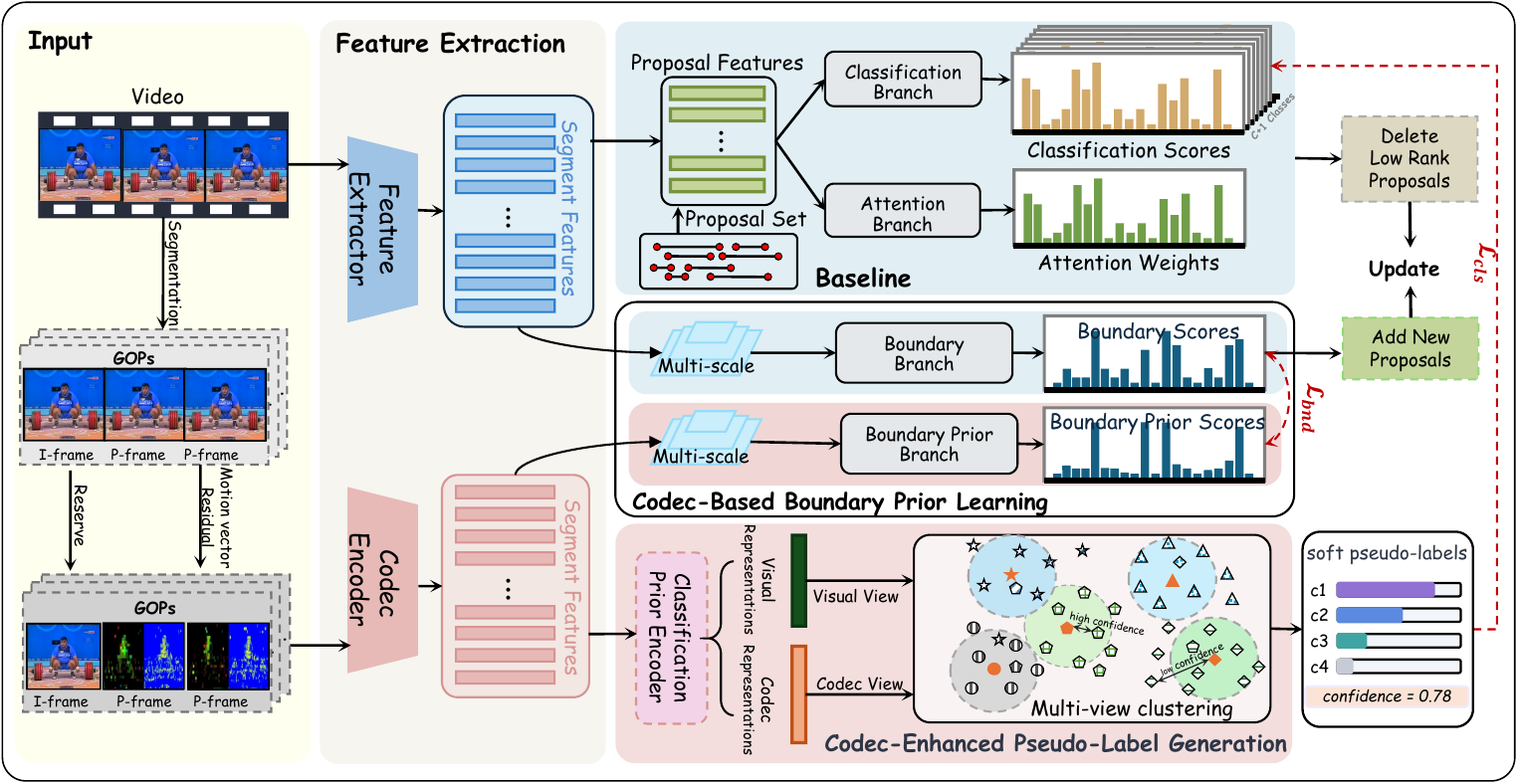}
  \caption{Overview of the proposed CoDy. CoDy executes from two views of visual and complementary codec view and is composed of P-MIL-based baseline, codec-enhanced pseudo-label generation and codec-based boundary prior learning.}
  \label{fig:method}
\end{figure*}



Video codecs exploit temporal redundancy by organizing a video into GOP. 
Each GOP begins with an intra-coded frame (I-frame), which serves as an appearance reference, followed by predictive frames (P-frames) encoded by motion vectors and residuals. 
Motion vectors describe the block-wise displacements selected for prediction, while residuals retain the visual changes that cannot be explained by the predicted motion. To obtain temporally aligned codec observations, we re-encode each video into fixed-length $0.5$-second GOPs. The $t$-th GOP is represented as:
\begin{equation} 
\mathcal{G}_{t} = \left( I_t, \{(\mathbf{M}_{t,j},\mathbf{R}_{t,j})\}_{j=1}^{n_t} \right),
\label{eq:gop}
\end{equation}
where $I_t$ is the I-frame and $\mathbf{M}_{t,j}$ and $\mathbf{R}_{t,j}$ are the motion-vector and residual fields of the $j$-th P-frame. 
Each GOP therefore yields a codec observation at a temporal resolution of $0.5$ seconds, while preserving both the spatial structure and temporal evolution of the compressed-domain signals. 
As summarized in Figure~\ref{fig:sec3}, we evaluate whether these predictive dynamics provide useful cues for the two core stages of UTAL including category discovery and boundary localization.
The following two observations evaluate these properties on THUMOS14. Detailed methodology and implementation details are provided in Appendix~A.

\subsection{Observation 1: Codec Dynamics Complement Category Discrimination}

We examine whether codec information provides a useful complementary view for category discrimination. We conduct a controlled comparison using the same videos, clustering algorithm, number of clusters, and random seed.
As illustrated in Figure \ref{fig:sec3}(a), we compare action category clustering results of video-level RGB and optical-flow descriptors with their codec-augmented counterparts.
The ground-truth labels are used to evaluate the quality of the generated pseudo-labels.
Without iterative refinement, codec augmentation view increases pseudo-label accuracy from $60.0\%$ to $63.5\%$ and macro-F1 from $52.87\%$ to $56.16\%$. It also reduces confusion between visually similar actions, particularly \texttt{JavelinThrow} and \texttt{ThrowDiscus}.
This improvement can be attributed to the complementary information retained by the codec. RGB features mainly describe objects and scenes, while optical flow features describe apparent pixel displacement. In contrast, motion vectors encode the block-wise motion selected during video compression, and residuals record changes that cannot be explained by motion compensation. 
These results show that codec dynamics boosts category clustering and the reliability of initial pseudo-labels.

\subsection{Observation 2: Codec Dynamics Exhibit Boundary Sensitivity}
We investigate whether codec dynamics respond more strongly to action boundaries than RGB and flow. For each $0.5$-second segment, we compute the normalized entropy of the RGB color distribution and the mean optical-flow magnitude as view-specific responses. The codec response aggregates the motion-vector magnitude and residual energy within each GOP:
\begin{equation}
M_t = \mathbb{E}\!\left[\|\mathbf{M}_t\|_2\right], \qquad
R_t = \mathbb{E}\!\left[\|\mathbf{R}_t\|_2^2\right],
\end{equation}
where $\mathbb{E}$ denotes averaging over all spatial locations and P-frames in a GOP, $M_t$ measures motion intensity and $R_t$ captures changes unexplained by motion compensation. The codec-based quantization response is computed as:
\begin{equation}
q_t^{c} =
\mathcal{N}\!\left(\lambda_m M_t+\lambda_r R_t+\lambda_g \sqrt{M_t,R_t}\right),
\label{eq:codec_quantization}
\end{equation}
where $\mathcal{N}(\cdot)$ is per-video normalization, and $\lambda_m$, $\lambda_r$ and $\lambda_g$ balance the two components. 
All responses are independently normalized, lightly smoothed, and aligned on the same temporal grid. Boundary annotations are used only for evaluation and do not participate in response construction.
As shown in Figure~\ref{fig:sec3}(b), the codec response exhibits sharper and more consistent elevations around action boundaries than the RGB and flow responses. 
Their stronger boundary alignment demonstrates that codec dynamics provide a more sensitive, class-agnostic cue for action-transition localization.

\section{Method}
\label{sec:method}
Let $\mathcal{D}=\{V_i\}_{i=1}^{N}$ denote a collection of unlabeled, untrimmed videos, and let $C$ be the number of latent action categories. 
Each video $V_i$ is represented by concatenated RGB and flow I3D features
$\mathbf{X}_i\in\mathbb{R}^{T_i\times D}$~\cite{carreira2017i3d}, together with its codec observations
$\{\mathcal{G}_{i,t}\}_{t=1}^{T_i}$ defined in Eq.~\eqref{eq:gop}. Given these inputs, the model predicts a set of action instances $\mathcal{P}=\left\{(s_i,e_i,c_i,q_i)\right\}_{i=1}^{K}$,
where $s_i$ and $e_i$ denote the temporal boundaries, $c_i\in\{1,\ldots,C\}$ is the discovered action category, and $q_i$ is the confidence score.


\subsection{Framework Overview}
\label{sec:overview}
As illustrated in Figure~\ref{fig:method}, CoDy builds upon the proposal-based localization framework P-MIL~\cite{ren2023pmil}.
Each candidate proposal is encoded into a proposal-level representation, from which a classification branch predicts latent category probabilities and an attention branch measures proposal attention weight. 
On this baseline, CoDy introduces two codec-guided branches. First, codec-enhanced pseudo-label generation aggregates GOP features into a video-level codec-semantic view. 
It jointly clusters them with RGB and flow representations to generate soft pseudo-labels, which are iteratively refined to supervise latent category discovery. 
Second, codec-based boundary prior learning exploits GOP-level motion-vector and residual dynamics to construct a class-agnostic boundary prior, which initializes the candidate proposal set and supervises an RGB and flow multi-scale boundary branch. 
During training, proposal scores remove low-confidence candidates, while the learned boundary responses introduce new proposals, enabling iterative refinement of category predictions and temporal boundaries.

\subsection{Codec-Enhanced Pseudo-Label Generation}
CoDy constructs a video-level codec view using a codec encoder inspired from CoPE-VideoLM~\cite{sarkar2026cope}. For the $t$-th GOP, the I-frame $I_t$ is processed by an image encoder to produce eight appearance tokens $\mathbf{U}_{t}^{I}\in\mathbb{R}^{8\times d}$. For each P-frame $j$, two lightweight encoders separately map its motion-vector and residual fields to four tokens, denoted by $\mathbf{U}_{t,j}^{M}\in\mathbb{R}^{4\times d}, \mathbf{U}_{t,j}^{R}\in\mathbb{R}^{4\times d}$. 
The I-frame tokens provide an appearance anchor, while the motion-vector and residual tokens describe encoded motion and prediction errors. We concatenate them to form the representation of GOP $t$: 
\begin{equation} 
\mathbf{G}_t = \left[ \mathbf{U}_{t}^{I}; \mathbf{U}_{t,1}^{M}; \mathbf{U}_{t,1}^{R}; \ldots; \mathbf{U}_{t,n_t}^{M}; \mathbf{U}_{t,n_t}^{R} \right]. 
\label{eq:gop_tokens} 
\end{equation}
Thus, a GOP containing $n_t$ P-frames is represented by $8+8n_t$ codec tokens.
After adding token-type and temporal positional embeddings, the GOP sequence is directly fed into a temporal Transformer.
A learnable video token aggregates the sequence, yielding the video-level codec representation:
\begin{equation} 
\mathbf{z}_i^{c} =
\mathrm{Transformer}\!\left(
[\mathrm{CLS}], \mathbf{G}_{i,1},\ldots,\mathbf{G}_{i,T_i}
\right)_{\mathrm{CLS}}. 
\label{eq:codec_video_feature} 
\end{equation}


To construct the visual view, CoDy obtains proposal-level features and attention weights from the baseline model. For video $i$, let $\mathbf{h}_{ir}$ and $a_{ir}$ denote the feature and attention weight of its $r$-th proposal, respectively. The proposal features are aggregated according to the attention weights to form a single video-level visual representation:
\begin{equation}
\mathbf{z}_i^{v} =
\sum_{r=1}^{R_i} a_{ir}\mathbf{h}_{ir},
\qquad
\sum_{r=1}^{R_i}a_{ir}=1,
\label{eq:visual_representation}
\end{equation}
where $R_i$ is the number of proposals in video $i$. This aggregation ensures that proposals more likely to contain actions contribute more strongly to the visual representation. The visual representation $\mathbf{z}_i^{v}$ and the codec representation $\mathbf{z}_i^{c}$ are independently normalized and used as two views for category clustering. While the visual view captures high-level appearance and motion semantics, the codec view provides compressed-domain motion and prediction information.

Then, CoDy performs confidence-aware consensus clustering over the two normalized views. For each video, $\mathbf{z}_i^{v}$ and $\mathbf{z}_i^{c}$ independently compute similarities to the cluster centers, producing view-specific soft assignments $\mathbf{p}_i^v$ and $\mathbf{p}_i^c$. The confidence of each view is determined by the $\ell_2$ distance between its feature and the nearest cluster center. The final pseudo-label $\mathbf{p}_i$ is obtained by confidence-weighted fusion of the two assignments:
\begin{equation}
\mathbf{p}_i
=
\alpha_i^v\mathbf{p}_i^v+
\alpha_i^c\mathbf{p}_i^c,
\qquad
\alpha_i^v+\alpha_i^c=1,
\label{eq:confidence_consensus}
\end{equation}
where the confidence weight for view $m\in\{v,c\}$ is:
\begin{equation}
\alpha_i^m = \frac{\exp\left(-\min_k d_{i,k}^m\right)}
{\exp\left(-\min_k d_{i,k}^v\right)
+\exp\left(-\min_k d_{i,k}^c\right)},
\label{eq:view_confidence}
\end{equation}
where $d_{i,k}^m$ denotes the $\ell_2$-norm distance between the feature of video $i$ in view $m$ and the $k$-th cluster center. Thus, the view whose feature is closer to a cluster center contributes more strongly to the final pseudo-label. The resulting consensus distribution $\mathbf{p}_i$ is used as a soft pseudo-label for classification learning and is periodically updated as visual and codec representations evolve during training. See Appendices~B and~C for details and clustering experiments.

\subsection{Codec-Based Boundary Prior Learning}
\label{sec:boundary_learning}
Motivated by the boundary sensitivity observed, CoDy converts codec dynamics into a class-agnostic boundary prior. Based on this observation, CoDy uses codec information as a class-agnostic boundary prior.
Specifically, CoDy employs two parallel multi-scale temporal convolutional networks with independent parameters. The visual boundary branch $H_v$ processes the snippet-level visual features $\mathbf{X}_i$, while the codec prior network $H_c$ processes the GOP-level features $\mathbf{G}_i$. Both networks use multi-scale temporal receptive fields to model both local dynamics and long-range temporal dependencies. The outputs are:
\begin{equation} 
\widehat{\mathbf{b}}_i^{v} = \sigma\!\left(H_v(\mathbf{X}_i)\right), \qquad \mathbf{b}_i^{c} = \mathcal{A}\!\left( \sigma\!\left(H_c(\mathbf{G}_i)\right) \right), 
\label{eq:visual_codec_boundary} 
\end{equation} 
where $\widehat{\mathbf{b}}_i^{v}\in[0,1]^{T_i}$ denotes the snippet-level visual boundary prediction, $\mathcal{A}$ aligns the codec prediction $\mathbf{b}_i^{c}$ with the visual temporal resolution and $\sigma$ is an activation function.



In addition to supervising the boundary branch, CoDy uses the temporally quantified codec response to initialize the proposal set. Our analysis shows that thresholding codec cues produces higher-quality candidates than applying the same procedure to RGB or flow. We therefore select high-response timestamps as candidate boundaries, pair them under duration constraints, and use the resulting intervals as the initial proposal instances. This codec-guided initialization enables CoDy to reduce unreliable candidates in early training, accelerating subsequent proposal selection and refinement.


\subsection{Training and Inference}
\label{sec:train_infer}

\paragraph{Iterative proposal refinement.} For a proposal $p=(s_p,e_p)$, CoDy re-extracts its proposal representation and obtains a category score $s_{p,c}^{\mathrm{cls}}$ and an attention score $s_p^{\mathrm{att}}$. 
Its endpoint consistency with the learned visual boundary sequence is measured by: 
\begin{equation} 
s_p^{\mathrm{bnd}} = \sqrt{ \max_{t\in\mathcal{N}(s_p)}\widehat{b}_{t}^{v} \; \max_{t\in\mathcal{N}(e_p)}\widehat{b}_{t}^{v} },
\label{eq:proposal_boundary_score} 
\end{equation} 
where $\mathcal{N}(u)=\{t\mid |t-u|\leq\delta\}$ denotes a local temporal neighborhood centered at endpoint $u$, and $\delta$ is a fixed tolerance radius. Taking the maximum response within each neighborhood accommodates minor temporal misalignment between the predicted boundary sequence and the proposal endpoints. The proposal ranking score is:
\begin{equation} 
S_p = \left(s_p^{\mathrm{cls}}+\epsilon\right)^{\alpha_{cls}} \left(s_p^{\mathrm{att}}+\epsilon\right)^{\alpha_{att}} \left(s_p^{\mathrm{bnd}}+\epsilon\right)^{\alpha_{bnd}}, 
\label{eq:proposal_score} 
\end{equation} 
where $\alpha_{cls}$, $\alpha_{att}$, and $\alpha_{bnd}$ control the contributions of category evidence, proposal foregroundness, and boundary consistency. At each refinement round $r$, CoDy removes low-scoring proposals with scores below $\tau$ and generates new proposals by pairing high-confidence local maxima of the boundary sequence under duration constraints: 
\begin{equation}
\mathcal{P}^{(r+1)} = \operatorname{Keep}\!\left(\mathcal{P}^{(r)}, \tau\right)
\cup
\operatorname{Generate}\!\left(\widehat{\mathbf{b}}^{v}\right).
\label{eq:iterative_proposal_update}
\end{equation}
The updated proposals are re-encoded and rescored in the next round, allowing category confidence and temporal extent to be refined jointly.
The same ranking and augmentation rule is used during training and inference. Further details and threshold experiments are provided in Appendices~B and~C.

\begin{table*}[t]
  \centering
  \setlength{\tabcolsep}{4pt}
  \fontsize{8.8pt}{10pt}\selectfont
  \begin{tabularx}{\textwidth}{c|c|CCCCCCC|CCC}
  \toprule
  \multirow{2}{*}{\textbf{Sup.}} &
  \multirow{2}{*}{\textbf{Method}} &
  \multicolumn{7}{c|}{mAP@IoU (\%)} &
  \multicolumn{3}{c}{AVG} \\
  \cmidrule(lr){3-9}\cmidrule(lr){10-12}
  & & 0.1 & 0.2 & 0.3 & 0.4 & 0.5 & 0.6 & 0.7
  & (0.1:0.5) & (0.3:0.7) & (0.1:0.7) \\
  \midrule

  \multirow{7}{*}{Weakly}
  & CoLA ~\cite{zhang2021cola}
  & 66.2 & 59.5 & 51.5 & 41.9 & 32.2 & 22.0 & 13.1
  & 50.3 & 32.1 & 40.9 \\
  & DGCNN ~\cite{shi2022dynamic}
  & 66.3 & 59.9 & 52.3 & 43.2 & 32.8 & 22.1 & 13.1
  & 50.9 & 32.7 & 41.3 \\
  & AUMN ~\cite{luo2021action}
  & 66.2 & 61.9 & 54.9 & 44.4 & 33.3 & 20.5 & 9.0
  & 52.1 & 32.4 & 41.5 \\
  & CSCL ~\cite{ji2021weakly}
  & 68.0 & 61.8 & 52.7 & 43.3 & 33.4 & 21.8 & 12.3
  & 51.8 & 32.7 & 41.9 \\
  & FAC-Net ~\cite{yu2023frame}
  & 67.6 & 62.1 & 52.6 & 44.3 & 33.4 & 22.5 & 12.7
  & 52.0 & 33.1 & 42.2 \\
  & ACGNet ~\cite{yang2022acgnet}
  & 68.1 & 62.6 & 53.1 & 44.6
  & 34.7 & 22.6 & 12.0
  & 52.6 & 33.4 & 42.5 \\
  & ACM-Net ~\cite{qu2021acmnet}
  & 68.9 & 62.7 & 55.0 & 44.6
  & 34.6 & 21.8 & 10.8
  & 53.2 & 33.4 & 42.6 \\

  \midrule

  \multirow{7}{*}{Unsup.}
  & STPN ~\cite{nguyen2018stpn}
  & 50.1 & 45.8 & 40.6 & 32.3 & 20.9 & 10.7 & 4.6
  & 37.9 & 21.8 & 29.3 \\


  & TSCN ~\cite{tang2020two}
  & 57.1 & 51.6 & 43.9 & 35.3 & 26.0 & 15.7 & 6.0
  & 42.8 & 25.4 & 33.7 \\


  & APSL ~\cite{liu2023apsl}
  & 57.7 & 52.4 & 44.1 & 35.9 & 27.9 & 18.5 & 10.0
  & 43.6 & 27.3 & 35.2 \\

  & UGCT ~\cite{yang2023ugct}
  & 63.4 & 57.8 & 51.7 & 44.0
  & 32.8 & 21.6 & 10.1
  & 49.9 & 32.0 & 40.2 \\

  & FEEL-V ~\cite{hu2026feel}
  & - & - & - & - & 29.3 & \underline{22.6} & 11.5
  & - & - & - \\

  & CLIP-AE ~\cite{xia2025clipae}
  & \textbf{65.1} & \underline{60.4} & \underline{52.7} & \underline{44.5}
  & \underline{33.2} & 22.3 & \underline{13.6}
  & \underline{51.2} & \underline{33.3} & \underline{41.7} \\

\cmidrule{2-12}

 \multicolumn{1}{c|}{\cellcolor{white}} & \textbf{CoDy (Ours)} & \underline{64.3} & \textbf{60.9} & \textbf{53.6} & \textbf{45.3} & \textbf{37.5} & \textbf{25.5} & \textbf{14.8}  & \textbf{52.3} & \textbf{35.3} & \textbf{43.1} \\

\multicolumn{1}{c|}{\cellcolor{white}} & \textit{\textbf{Improvement} ($\Delta$)} & (-0.8) & (+0.5) & (+0.9) & (+0.8) & (+4.3) & (+2.9) & (+1.2) & (+1.1) & (+2.0) & (+1.4) \\
\bottomrule
\end{tabularx}
  \caption{Comparison results on the THUMOS14. The best and second-best results are in \textbf{bold} and \underline{underlined}.}
  \label{tab:thumos}
\end{table*}

\begin{table}
  \centering
    \setlength{\tabcolsep}{2pt} \fontsize{8.8pt}{10pt}\selectfont 
    \begin{tabular}{c|cccc}
        \toprule
        \multirow{2}{*}{\textbf{Method}} & \multicolumn{4}{c}{mAP@IoU (\%)} \\ 
        \cmidrule(lr){2-5}
         & 0.5 & 0.75 & 0.95 & AVG \\ 
          \midrule
  STPN ~\cite{nguyen2018stpn}
  & 23.9 & 14.9 & 3.8 & 15.0 \\


  TSCN ~\cite{tang2020two}
  & 25.9 & 16.2 & 3.8 & 16.3 \\

  WSAL-BM + UGCT~\cite{yang2023ugct}
  & 30.1 & 19.2 & 3.9 & 18.5 \\

  TSCN + UGCT~\cite{yang2023ugct}
  & 30.3 & 19.4 & 4.3 & 18.7 \\

  STPN + UGCT~\cite{yang2023ugct}
  & \textbf{32.5} & \underline{20.3} & \underline{4.5} & \underline{19.8} \\

  \midrule

  \textbf{CoDy (Ours)}
  & \underline{31.9} & \textbf{20.3} & \textbf{5.0} & \textbf{20.2} \\

  \textit{\textbf{Improvement} ($\Delta$)}
  & (-0.6)
  & (+0.0)
  & (+0.5)
  & (+0.4)\\

    \bottomrule
\end{tabular}
\caption{Comparison results on the ActivityNet v1.3. The best and second-best results are in \textbf{bold} and \underline{underlined}.}
    \label{tab:anet}
\end{table}

\paragraph{Training.} 
During training, CoDy uses soft pseudo-labels to supervise the classification and attention branch. Let $\mathbf{p}_i$ and $\widehat{\mathbf{p}}_i$ denote the soft pseudo-label and prediction of video $i$. The classification loss is:
\begin{equation}
\mathcal{L}_{\mathrm{cls}} = -\frac{1}{N}
\sum_{i=1}^{N}
\sum_{c=1}^{C}
\mathbf{p}_{i,c}\log \widehat{\mathbf{p}}_{i,c}.
\label{eq:classification_loss}
\end{equation}
For boundary learning, we jointly optimize the visual boundary branch and codec prior boundary branch:
\begin{equation}
\mathcal{L}_{\mathrm{bnd}} = \frac{1}{N}
\sum_{i=1}^{N}
\left[
D\!\left(
\widehat{\mathbf{b}}_i^{v},
\mathbf{b}_i^{c}
\right) + \lambda_{c}\,
D\!\left(\mathbf{b}_i^{c}, q_t^{c}\right)
\right],
\label{eq:boundary_loss}
\end{equation}
where $D(\cdot,\cdot)$ denotes the mean squared error. $\mathbf{b}_i^{c}$ is anchored to the temporally codec response $q_t^{c}$ derived in Eq.~\eqref{eq:codec_quantization}, which provides a fixed, non-learnable prior. $\lambda_c$ controls the anchor strength and is gradually decayed with training depth. The complete objective is: 
\begin{equation} 
\mathcal{L}_{\mathrm{total}} = \mathcal{L}_{\mathrm{cls}} + \lambda \mathcal{L}_{\mathrm{bnd}}, 
\label{eq:joint_training_loss} 
\end{equation} 
where $\lambda$ is the balance parameter.
\paragraph{Inference.} At inference, CoDy discards pseudo-label generation and the codec prior branch, retaining only the P-MIL backbone and the learned visual boundary branch.
Without ground-truth labels or codec temporal priors, the model performs a fixed number of self-refinement rounds. 
Starting from the baseline proposals, CoDy iteratively updates the proposal set using Eq.~\eqref{eq:iterative_proposal_update}.
At each round, it predicts proposal scores and a boundary sequence, and then updates the proposal set. 
The final detections are obtained after the last refinement round and standard score suppression.

\section{Experiments}
\label{sec:experiments}

\subsection{Experimental Setup}
\paragraph{Datasets.}
We evaluate CoDy on two standard temporal action localization benchmarks. THUMOS14~\cite{idrees2017thumos} contains 413 temporally annotated untrimmed videos across 20 action categories, including 200 validation and 213 test videos. 
Following standard protocols, we train on the validation set and evaluate on the test set. 
ActivityNet v1.3~\cite{heilbron2015activitynet} covers 200 activity categories across 19,994 untrimmed videos, divided into 10,024 training, 4,926 validation, and 5,044 test videos. We train on the training set and evaluate on the validation set.

\paragraph{Metrics.}
For THUMOS14, we report detection mAP at temporal IoU thresholds from $0.1$ to $0.7$ with a step of $0.1$, along with their average. For ActivityNet v1.3, we report mAP at $0.5$, $0.75$, $0.95$, and the average mAP over thresholds from $0.5$ to $0.95$ with a step of $0.05$. 
%

\paragraph{Implementation details.}
We adopt the UTAL variant of P-MIL~\cite{ren2023pmil} as baseline. Visual features are extracted using I3D models pretrained on Kinetics, where RGB and optical-flow features are concatenated to form $D=2048$-dimensional segment features. For codec processing, videos are sampled at 30 FPS and divided into 0.5-second groups, each consisting of 1 I-frame and 14 P-frames. We extract motion vectors and residuals from the H.264 bitstream, and set the codec token dimension to $d=64$. Both boundary branches use four-block multi-scale TCNs with dilation rates of $\{1,2,4,8\}$ and a kernel size of $3$. 
The model is trained for 200 epochs using Adam with a learning rate of $5\times10^{-5}$, a batch size of $10$, and a weight decay of $10^{-3}$. We set $\tau = 0.3$, $\lambda = 1$ and initialize $\lambda_c = 1$ in all experiments. More implementation and training details are provided in Appendix~B. A comprehensive analysis of the computational cost of CoDy is presented in Appendix~C.

\subsection{Main Results}
\label{sec:main_results}
Table~\ref{tab:thumos} and Table~\ref{tab:anet} summarize the results on THUMOS14 and ActivityNet v1.3. CoDy consistently outperforms existing unsupervised methods at most IoU thresholds and achieves the strongest overall accuracy. The performance gains are particularly pronounced at stricter IoU thresholds, demonstrating that codec-prior dynamics effectively enhance temporal boundary precision. The slightly lower performance at the lowest threshold suggests that codec priors may mistake scene changes and camera motion for action transitions.
The consistent gains across both datasets demonstrate the broad applicability of codec cues in UTAL. On densely annotated THUMOS14, codec-based temporal variations resolve closely spaced action boundaries. On ActivityNet v1.3, which features longer and sparser activities, the same cues improve localization precision. These findings confirm that CoDy successfully leverages codec information as a powerful prior for proposal refinement. 



\subsection{Ablation Study}
\label{sec:ablations}
\subsubsection{Impact of Key Components.}   
We conduct an ablation study to evaluate the two core components of CoDy: codec-enhanced pseudo-label generation and codec-prior boundary guidance. As summarized in Table~\ref{tab:core}, the \textit{Classification} setup uses the codec video-level representation as a complementary view during multi-view clustering, while the \textit{Localization} setup uses the codec-based boundary prior to supervise the boundary branch. Both components consistently improve performance over the codec-free baseline. Specifically, the semantic view enhances pseudo-label quality for classification, whereas the boundary prior improves proposal temporal precision. Combining both components yields the highest accuracy, outperforming either standalone branch.



\begin{table}
  \centering

    \setlength{\tabcolsep}{5pt} \fontsize{8.8pt}{10pt}\selectfont
    \begin{tabular}{cc|ccccccc}
        \toprule 
        \multicolumn{2}{c|}{\textbf{Main components}} & \multicolumn{4}{c}{mAP@IoU (\%)} \\ 
        \midrule
      Classification   & Localization & 0.3 & 0.5 & 0.7 & AVG \\ 
        \midrule
  &    & 44.8 & 30.8 & 8.5 & 35.7 \\
 & \ding{51}  &  48.8 & 34.9 & 11.0 & 38.4 \\
\ding{51} &  &  49.9 & 35.2 & 12.1 & 39.5 \\
  
\ding{51} & \ding{51} &  \textbf{53.6} & \textbf{37.5} & \textbf{14.8} & \textbf{43.1} \\

        \bottomrule
    \end{tabular}
        \caption{Ablation study of key components on THUMOS14. } 
            \label{tab:core}
\end{table}

\subsubsection{Impact of Proposal Sources.}

\begin{figure}[t]
    \centering
    \begin{subfigure}{0.49\columnwidth}
        \centering
        \includegraphics[width=\linewidth]{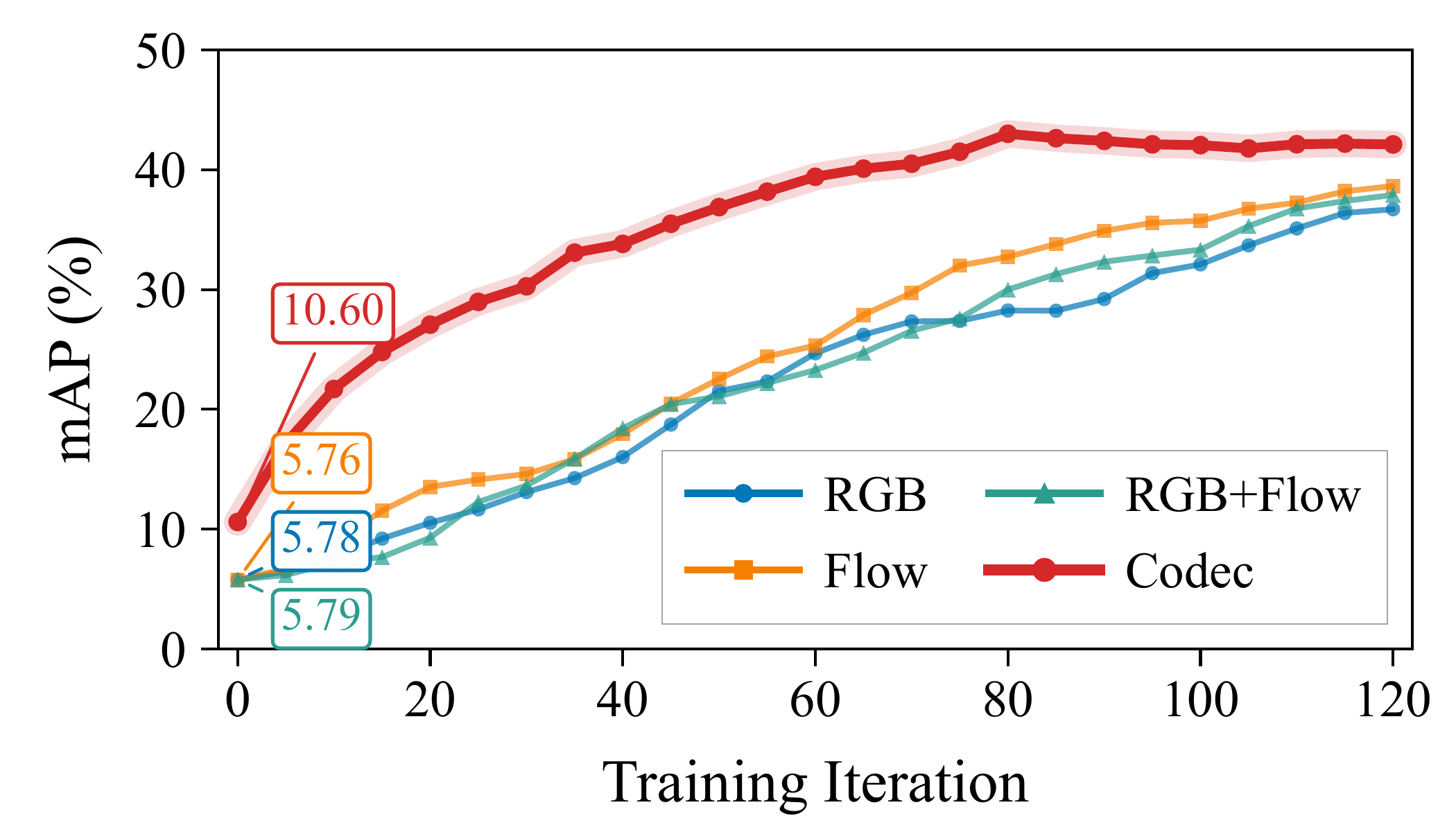}
        \caption{Proposal sources}
        \label{fig:fig4}
    \end{subfigure}
    \hfill
    \begin{subfigure}{0.49\columnwidth}
        \centering
        \includegraphics[width=\linewidth]{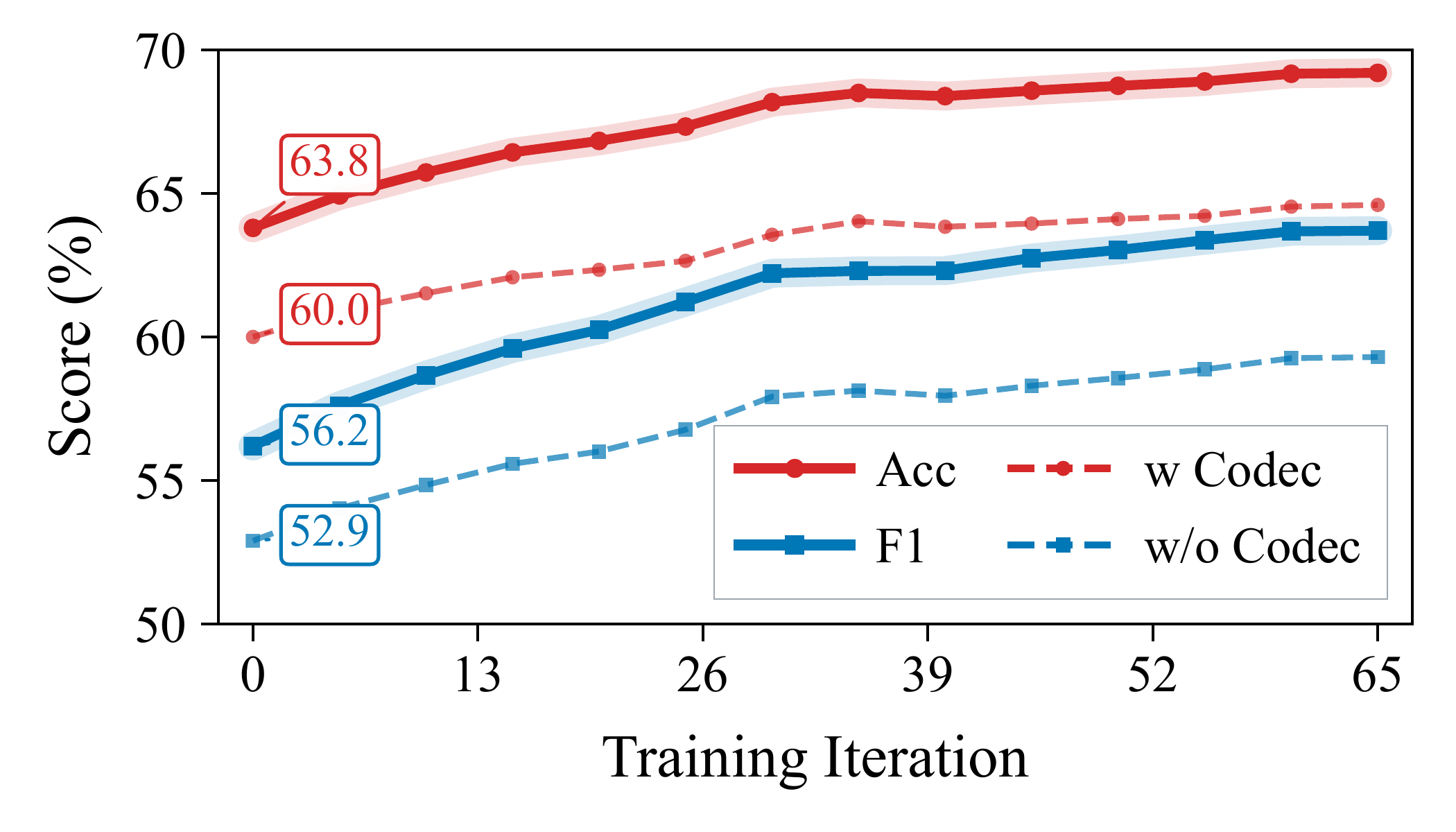}
        \caption{Pseudo-label quality}
        \label{fig:fig5}
    \end{subfigure}
    \caption{Ablation study of codec for proposal source and pseudo-label quality.}
    \label{fig:iter}
\end{figure}

To evaluate how different signal sources affect candidate
initialization, we test class-agnostic proposals built from various features over iterative refinement. Video-level category labels are provided to all variants solely to eliminate classification bias and isolate proposal quality. 
As shown in Figure~\ref{fig:iter}(a), initialization sources exhibit distinct trajectories. At iteration 0, codec proposals achieve $\textbf{10.60\%}$ average mAP, markedly outperforming RGB at $5.78\%$, flow at $5.76\%$, and their fusion at $5.79\%$. This initial lead proves that codec signals supply richer and sharper temporal boundary cues.
This superior initialization accelerates convergence and maintains higher overall accuracy throughout training. Conversely, non-codec variants yield inferior trajectories similar to random initialization.

\subsubsection{Impact of Codec View for Pseudo-Label Quality.}

We compare two pseudo-label generation settings under iterative training, one using RGB and flow visual views alone and the other incorporating the codec view as a complementary clustering signal. At every pseudo-label refresh, we evaluate the generated labels using aligned accuracy and macro-F1. After each iteration, the Hungarian algorithm assigns the discovered clusters to category labels. Ground truth video level labels are available only to the evaluator.
As shown in Figure~\ref{fig:iter}(b), even before model iteration at iteration 0, incorporating the codec view directly produces more accurate and more balanced pseudo-labels. As training proceeds, visual and codec representations are repeatedly updated, while CoDy maintains its pseudo-label quality advantage across successive refreshes. This shows that codec representations provide complementary information to RGB and flow.



\subsubsection{Impact of Motion Vector and Residual Features.}
\begin{table}
    \centering
    \setlength{\tabcolsep}{5pt}
    \fontsize{8.8pt}{10pt}\selectfont
    \begin{tabular}{c|c|cccc}
        \toprule
        \multirow{2}{*}{\textbf{}} 
        & \multirow{2}{*}{\textbf{Codec Feature}} 
        & \multicolumn{4}{c}{mAP@IoU (\%)} \\
        \cmidrule(lr){3-6}
        & & 0.3 & 0.5 & 0.7 & AVG \\
        \midrule
        
        \multirow{2}{*}{Classification}
        & Residual-only 
        & 48.9 & 33.6 & 10.1 & 38.9 \\
        & MV-only 
        & 49.6 & 35.5 & 11.1 & 40.9 \\
        
        \midrule
        
        \multirow{2}{*}{Localization}
        & Residual-only 
        & 50.6 & 35.7 & 12.4 & 41.8 \\
        & MV-only 
        & 49.5 & 34.6 & 11.8 & 39.8 \\
        
        \midrule
        \textbf{CoDy(Ours) }
        & MV+Residual 
        & \textbf{53.6} & \textbf{37.5} & \textbf{14.8} & \textbf{43.1} \\
        
        \bottomrule
    \end{tabular}
        \caption{Ablation study of motion vector and residual features on classification and localization performance.}
            \label{tab:variable_comparison}
\end{table}

We compare MVs, residuals, and their fusion as codec representations to quantify their respective contributions. As shown in Table~\ref{tab:variable_comparison}, MVs are more effective for the classification branch, as they provide explicit motion cues that facilitate semantic understanding. Residuals perform better for boundary modeling, since they preserve local appearance variations and temporal discontinuities around action transitions. Combining MVs and residuals, CoDy achieves the best performance.

\section{Conclusion}
We present CoDy, a codec-guided framework that exploits complementary information from the video bitstream for UTAL. CoDy integrates a codec view with RGB and flow representations to improve latent category discovery and iterative pseudo-label generation. 
Meanwhile, motion-vector and residual dynamics provide class-agnostic priors for visual boundary learning and proposal refinement. Extensive experiments on THUMOS14 and ActivityNet v1.3 demonstrate the superior performance of CoDy. These results establish codec dynamics as a complementary source of supervision for learning action categories and temporal boundaries.

{\small

}

\end{document}